%% file: output.tex
\documentclass[10pt,twocolumn,letterpaper]{article}
\usepackage{cvpr}  

\def\paperID{561} 
\def\confName{CVPR}
\def\confYear{2025}
\usepackage[ruled,vlined]{algorithm2e}
\usepackage{graphicx}	
\usepackage{amsmath}	
\usepackage{amssymb}	
\usepackage{booktabs}
\usepackage{times}
\usepackage{microtype}
\usepackage{epsfig}
\usepackage{float}
\usepackage{placeins}
\usepackage{color, colortbl}
\usepackage{stfloats}
\usepackage{enumitem}
\usepackage{tabularx}
\usepackage{xstring}
\usepackage{multirow}
\usepackage{xspace}
\usepackage{url}
\usepackage{subcaption}
\usepackage{hyperref}

\title{Stacking Brick by Brick: Aligned Feature Isolation for \\Incremental Face Forgery Detection}
\author{Jikang Cheng$^1$\thanks{Equal contribution}, Zhiyuan Yan$^{2}$$^{ *}$, Ying Zhang$^3$,  Li Hao$^2$, Jiaxin Ai$^1$, Qin Zou$^1$, Chen Li$^3$, Zhongyuan Wang$^1$\thanks{Corresponding author} \\
	School of Computer Science, Wuhan University$^1$\\
        School of Electronic and Computer Engineering, Peking University Shenzhen Graduate School$^2$\\
        WeChat Vision, Tencent Inc.$^3$\\
	{\tt\small ChengJiKang@whu.edu.cn}}

\begin{document}
\maketitle
\begin{abstract}
The rapid advancement of face forgery techniques has introduced a growing variety of forgeries.
Incremental Face Forgery Detection (IFFD), involving
gradually adding new forgery data to fine-tune the previously trained model, has been introduced as a promising strategy to deal with evolving forgery methods.
However, a naively trained IFFD model is prone to catastrophic forgetting when new forgeries are integrated, as treating all forgeries as a single ``Fake" class in the Real/Fake classification can cause different forgery types overriding one another, thereby resulting in the forgetting of unique characteristics from earlier tasks and limiting the model's effectiveness in learning forgery specificity and generality.
In this paper, we propose to stack the latent feature distributions of previous and new tasks brick by brick, \textit{i.e.},  achieving \textbf{aligned feature isolation}. 
In this manner, we aim to preserve learned forgery information and accumulate new knowledge by minimizing distribution overriding, thereby mitigating catastrophic forgetting.
To achieve this, we first introduce Sparse Uniform Replay (SUR) to obtain the representative subsets that could be treated as the uniformly sparse versions of the previous global distributions.
We then propose a Latent-space Incremental Detector (LID) that leverages SUR data to isolate and align distributions. 
For evaluation, we construct a more advanced and comprehensive benchmark tailored for IFFD. 
The leading experimental results validate the superiority of our method. Code is available at \textit{\href{https://github.com/beautyremain/SUR-LID}{https://github.com/beautyremain/SUR-LID}}.
\end{abstract}
\begin{figure}[t]
    \centering
    \includegraphics[width=1\linewidth]{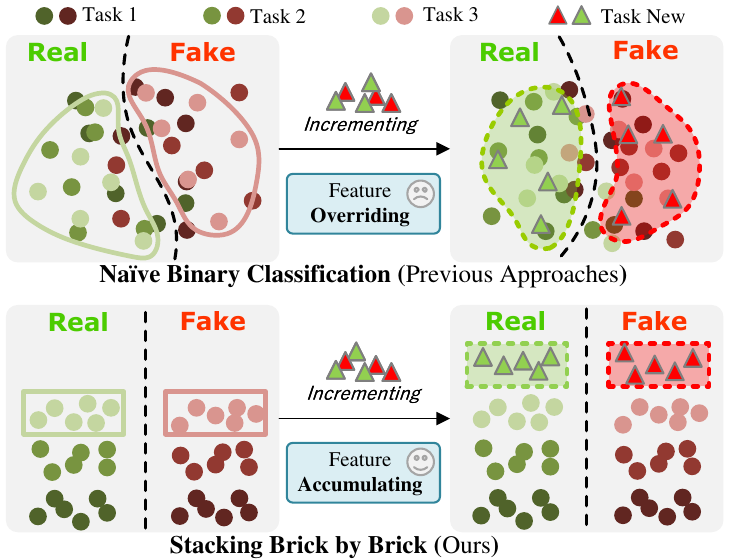}
    \caption{Illustration of the proposed aligned feature isolation in the latent space. 
    Previous approaches (top) typically treat all forgeries, both old and new, as a single ``Fake" class during incremental learning, causing feature distributions to override each other and limiting their ability to learn forgery specificity and generality.
    In contrast, we (bottom) propose \textbf{incrementally} adding new task distributions with \textbf{isolation and alignment}, akin to stacking new tasks ``\textit{brick by brick}" to the previous ones in the latent space.
    See Fig.~\ref{fig:Umap-dist} for the experimental results of latent space distribution.}
    \vspace{-0.2cm}
\label{fig:first_impression}
\end{figure}
\section{Introduction}

The rise of face forgery techniques poses substantial threats to society, drawing increased attention from researchers who study the risks of misuse, particularly in identity theft, misinformation, and violations of privacy. Hence, developing effective detection methods is essential to safeguard personal security and maintain public trust in digital interactions. Existing methods~\cite{ucf,ed4,huang2023implicit,SBI,prodet} predominantly focus on training a generalized face forgery detector with a limited number of training data. 
However, given the ever-increasing diversity of face forgery techniques in the real world, it is somewhat idealistic to expect a generalized model to effectively detect all types of forgery solely relying on limited training data~\cite{prodet}. Concurrently, training a new model with all available data whenever a new forgery emerges can lead to significant issues of computational expenses, storage limitations, and privacy implications.
Therefore, adopting an incremental learning research paradigm for face forgery detection could address a wider range of application scenarios considering the ever-increasing volume of forgery data.

To date, only a few methods have explored the field of Incremental Face Forgery Detection (IFFD)~\cite{cored,dfil,hdp,dmp}. These methods propose to preserve representative information from previous tasks via various replay strategies, such as selecting center and hard samples~\cite{dfil}, generating representative adversarial perturbations~\cite{hdp}, and considering mixed prototypes~\cite{dmp}. 
However, since IFFD consistently aims at learning the same simple binary classification, the backbone extractor is more prone to casually override the \textit{global} feature distribution of the previous tasks with the new incrementing one. 
This situation makes the issue of catastrophic forgetting in IFFD particularly pronounced. Although current methods have proposed various strategies for replay and regularization, they primarily focus on preserving a few particular representative samples (such as center and hard samples in DFIL~\cite{dfil}) and maintaining their feature consistency. Consequently, they struggle to maintain and thus organize the \textit{global} feature distributions learned previously, thereby challenging to mitigate distribution overriding.

In this paper, we propose to stack feature distributions of previous and new tasks brick by brick in the latent space, \textit{i.e.}, achieving aligned feature isolation. As shown in Fig.~\ref{fig:first_impression}, we use the term ``brick" to describe our feature distributions because we force them to be mutually isolated rather than overridden    \footnote{Each distribution is not required to be strictly rectangular like ``brick''.}, while ``brick by brick'' refers to aligning the binary decision boundary of the incrementing task with all previous tasks one by one. The advantages of implementing the proposed ``stacking brick by brick'' are two-fold. Firstly, 
\textit{feature isolation} allows for reducing feature distribution override between new and previous domains and thus better preserving the knowledge acquired from previous tasks. Secondly, \textit{one-by-one decision alignment} ensures that the accumulated diverse forgery information can be effectively utilized for final binary face forgery detection during incremental learning.

To achieve aligned feature isolation, we propose a novel IFFD method called SUR-LID. Specifically, one prior requirement for aligning and isolating all feature distributions is to obtain replay subsets that could represent the previous global distributions. Therefore, we first propose a Sparse Uniform Replay (SUR) strategy that selects replay samples based on their stability and distribution density. The distribution of the SUR subset could be treated as a uniformly sparse version of the original global distribution.
With the distribution preserved by SUR, we can propose a Latent-space Incremental Detector (LID) to achieve aligned feature isolation. LID employs an isolation loss to isolate each distribution, which is enhanced by distribution re-filling that could further recover and simulate the previous global distribution based on SUR data. Then, incremental decision alignment is introduced to enforce the new task to have a decision boundary that is aligned with all previous ones. 
Additionally, we further introduce two carefully designed incrementing protocols to improve the experimental evaluation of IFFD performance. 
The leading results demonstrate the superiority of the proposed method.
Our contributions can be summarized as:
\begin{itemize}
    \item We propose to stack the feature distributions of the previous and new tasks brick by brick in the latent space, \textit{i.e.}, achieving aligned feature isolation. It could mitigate feature overriding and effectively accumulate learned diverse forgery information to improve face forgery detection.
    \item For aligned feature isolation, we introduce SUR to store previous global distribution and LID that leverages SUR data to achieve feature isolation and alignment.
    \item We carefully construct a new comprehensive benchmark for evaluating IFFD, which includes diverse latest forgery methods and two protocols corresponding to practical real-world applications.
\end{itemize}

\section{Background}
\subsection{Preliminary of IFFD}
\textbf{Training Paradigm.}
In incremental learning, new data is introduced sequentially to fine-tune a model that has already been trained on prior tasks, and the complete prior data remains inaccessible~\cite{para-isolation}. Compared to re-training a model from scratch with all available data, this paradigm allows incrementally leveraging new data with reduced computational overhead and storage demands.\\
\textbf{Research Objective.}
Following~\cite{hdp}, we aim to address the crucial issue of catastrophic forgetting in incremental learning. 
Namely, the model performance on previously learned tasks may degrade significantly when incrementing new tasks, that is, forgetting the learned knowledge.\\
\textbf{Replay Set.}
Replay set refers to storing a tiny subset of data from the learned training set. With minimal additional storage overhead, it could significantly improve the model ability to retain previously learned knowledge while also allowing design flexibility for enhanced incremental learning.
\subsection{Face Forgery Detectors}
The existing methods mostly focus on the generalization of the detector to deal with the severe threat strive from face forgery. For example, given the observed model bias in the detector, various methods~\cite{ucf,Disentangle,ed4} have been proposed to mitigate general model biases present in forgery samples. In latent space, there are also methods~\cite{prodet,lsda} investigating the feature organization and fusion to mine and diversify the forgery information for generalizable forgery detectors. These methods~\cite{xception,ucf,Disentangle,ed4,huang2023implicit,prodet,lsda,lips,spsl} are proposed to capture general forgery information from limited seen data and exhibit promising performance in a few unseen data. 

However, considering the rapid evolution of face forgery techniques, it is impractical to rely on limited seen data to train an ideal generalizable detector. Therefore, the paradigm of incremental learning could be a superior alternative to adapting diverse and evolving forgery techniques.
\subsection{Incremental Face Forgery Detection}

General incremental learning methods are widely investigated and can be categorized into parameter isolation~\cite{para-isolation}, parameter regularization~\cite{lwf,ewc,mas}, and data replay~\cite{icarl,scr}. Nonetheless, only a few approaches focus on building an effective framework for incremental face forgery detection. Among them, CoReD~\cite{cored} leverages distillation loss to maintain previous-task knowledge, whereas DFIL~\cite{dfil} enhances this by using both center and hard samples for replay. HDP~\cite{hdp} refines universal adversarial perturbations (UAP~\cite{uap}) as a replay mechanism for earlier task knowledge. DMP~\cite{dmp} creates a replay set using mixed prototypes to encapsulate previous tasks.

Despite the fact that existing methods replay and maintain the knowledge from few representative data (\textit{e.g.}, center and hard samples), they cannot maintain and organize the global distributions of previous and incrementing tasks. Consequently, the previous global distribution is often overridden by the incrementing one, thus leading to the forgetting issue and insufficient learning of forgery specificity and generality.
\section{Methodology} 
\subsection{Rationale Behind Aligned Feature Isolation}
During training, the backbone extractor learned to map the image-space input to the representative feature in the latent space (\textit{i.e.}, image-feature mapping).
Hence, the global distribution of the extracted features could reflect the knowledge learned by the backbone extractor from the training task. Consequently, overriding previous distributions could destroy the previously learned image-feature mapping, and thus forgetting knowledge from the previous tasks. Moreover, the latent-space organization is proven to be crucial for model effectiveness~\cite{latent_support1,latent_support2,prodet}. The existing methods~\cite{cored,dfil,hdp,dmp} that preserve a few representative data points could only maintain performance on these certain points instead of the global distribution. Meanwhile, it is also challenging to organize the latent space of previous and incrementing tasks without preserving global distribution. 

Therefore, we propose aligned feature isolation to improve IFFD with three steps: 1) Storing replay subsets that could represent global distribution rather than a limited number of particular points.
2) Isolating global distributions of each task to minimize override, thereby allowing for the incremental accumulation of increasingly diverse forgery information. 3) Leveraging the accumulated forgery information obtained from isolation via decision alignment, thus enhancing the final binary face forgery detection.


\begin{figure*}[ht]
    \centering
    \includegraphics[width=0.95\linewidth]{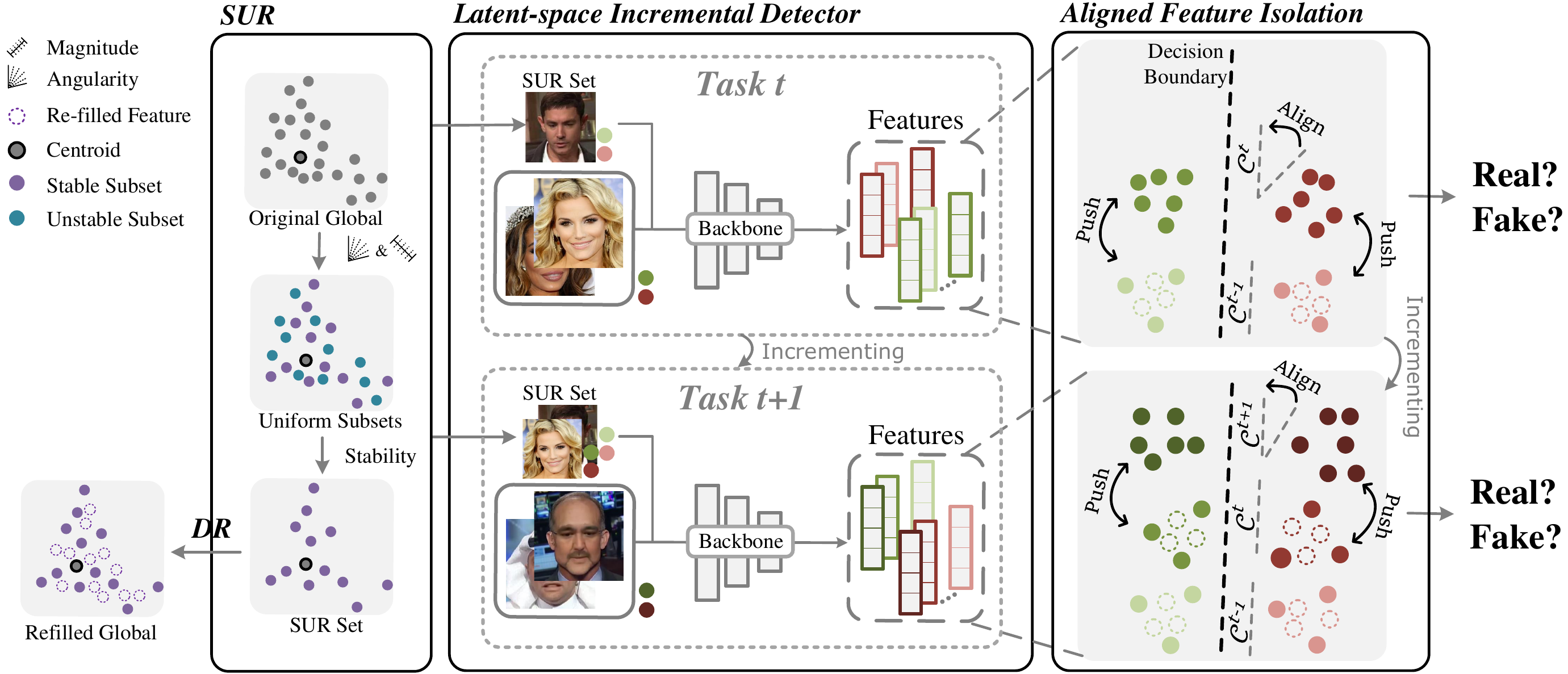}
    \caption{Overall framework of the proposed method.}
    \label{fig:pipeline}
    \vspace{-0.1cm}
\end{figure*}
\subsection{Overall Framework}
In this paper, the proposed aligned feature isolation for IFFD has two crucial components, that is, a replay strategy named Sparse Uniform Replay (SUR) and a detection model named Latent-space Incremental Detector (LID). We deploy SUR to store data after the training for one task is complete. Then, the SUR data is merged with the next training set to train the LID for incremental face forgery detection. The overall framework is shown in Fig.~\ref{fig:pipeline}. 
\begin{figure}[htbp]
    \centering
    \includegraphics[width=1\linewidth]{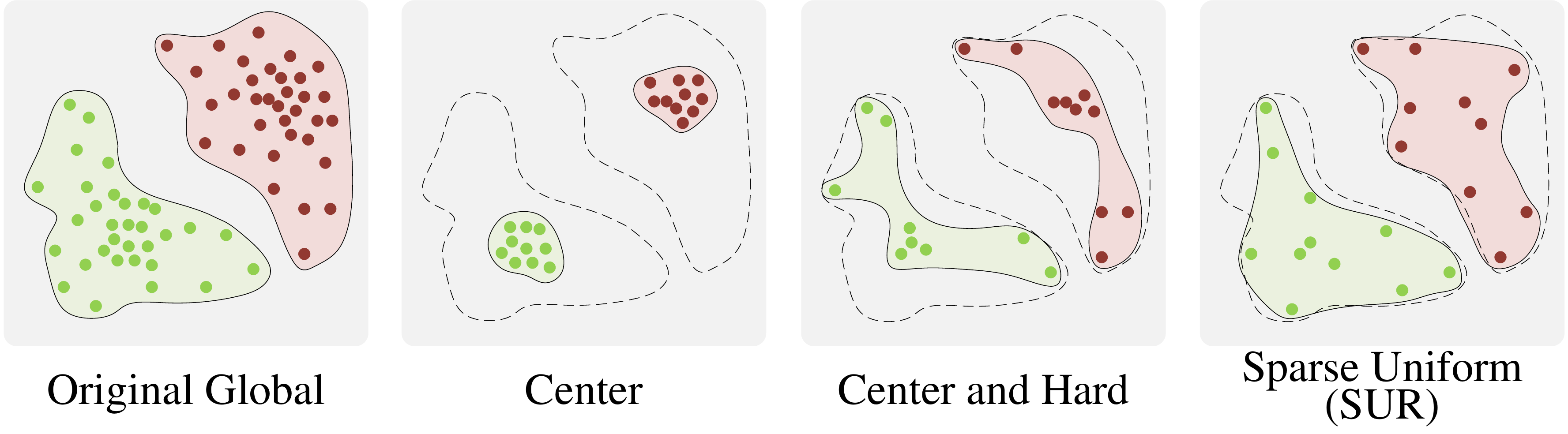}
    \caption{Illustration of different replay strategies. Using Center~\cite{cored,dmp} or Center and Hard~\cite{dfil} cannot preserve the global feature distribution, while the proposed SUR could uniformly sample a sparse version of the original global distribution.}
    \label{fig:replay-compare}
    \vspace{-0.5cm}
\end{figure}
\subsection{Sparse Uniform Replay (SUR)} 
To realize the proposed aligned feature isolation, a key prerequisite is having the reference of the previous $t$-th task global feature distributions 
when incrementing the new $(t+1)$-th task. 
Therefore, as shown in Fig.~\ref{fig:replay-compare}, we propose the Sparse Uniform Replay (SUR) strategy, which seeks to select \textit{stable} representations\footnote{Stable representation refers to the features that are being extracted uniformly when irrelevant content in input is altered~\cite{stable1,stable2}.} from the previous training set with \textit{high-dimensional uniformity} in the latent space. 
Specifically, maintaining uniformity in the replay set allows it to approximate the global distribution, rather than representing solely a localized region in the original distribution. Meanwhile, sampling the stably extracted features can reduce the risk of including abnormal outliers in the replay set.

Considering one task usually contains both real and fake domains, to simplify notation, we use $\mathbf{F}^t \in \mathbb{R}^{n \times d}$ and $\mathbf{X}^t \in \mathbb{R}^{n \times 3 \times w \times h}$ to denote one specific domain of features and their corresponding images, which could be either real or fake in $t$-th task, where $n$ is the number of sample, $d$ is the dimension of feature, $w$ and $h$ is the width and height of images.  
Given the trained backbone extractor of the $t$-th task $\mathcal{E}^t$, $\mathbf{F}^t$ could be generated by $\mathbf{F}^t=\mathcal{E}^t(\mathbf{X}^t)$. 
Firstly, we leverage centroids as the reference to uniformly sample the replay set,
which can be calculated as $\mathbf{c}^t = \text{avg}(\mathbf{F}^t) \in \mathbb{R}^{d}$. Sampling uniformly in the high-dimensional feature space requires considering both magnitude and angularity. Specifically, the \textbf{magnitude} from $\mathbf{c}^t$ to each feature in $\mathbf{F}^t$ can be written as: 
\begin{equation}
    \mathbf{M}^t = \|\mathbf{F}^t - \mathbf{c}^t\|_2,
\end{equation}
where $\|\ast\|_2$ represents calculating the Euclidean norm. Subsequently, the high-dimensional \textbf{angularity} matrix $\mathbf{A}^t$ can be calculated as: 
\begin{equation}
\mathbf{A}^t = \frac{(\mathbf{F}^t - \mathbf{c}^t)}{\|\mathbf{F}^t - \mathbf{c}^t\|_2}.
\end{equation}
Then, we leverage the shuffle consistency~\cite{core,ed4,dcl} to quantize the stability of the learned representation. Namely, since the forgery information is predominantly fine-grained and remains unaffected by shuffling, the forgery features should be consistent~\cite{core,ed4,dcl} with or without shuffling. Therefore, We conduct grid shuffle~\cite{shuffle} on $\mathbf{X}^t$ to generate $\tilde{\mathbf{X}}^t$ and thus obtain the features of shuffled data as $\tilde{\mathbf{F}}^t=\mathcal{E}^t(\tilde{\mathbf{X}}^t)$. Hence, the $i$-th element ($s^t_i$) in the \textbf{stability} matrix $\mathbf{S}^t$ is calculated using $i$-th features ($\tilde{\mathbf{f}}^t_i$ and $\mathbf{f}^t_i$) from $\tilde{\mathbf{F}}^t$ and $\mathbf{F}^t_i$ as: 
\begin{equation}
    s^t_i = \frac{\tilde{\mathbf{f}}^t_i \cdot (\mathbf{f}^t_i)^\text{T}}{\|\tilde{\mathbf{f}}^t_i\|_2 \cdot \|\mathbf{f}^t_i\|_2},
\end{equation}
where the superscript $\text{T}$ denotes the transpose matrix.
Intuitively, all three factors (\textit{i.e.}, $\mathbf{M}^t \in \mathbb{R}^{n} $, $\mathbf{A}^t\in \mathbb{R}^{n \times d}$, and $\mathbf{S}^t \in \mathbb{R}^{n}$) should be simultaneously considered to obtain uniform and stable representation. However, achieving an ideal strategy demands high-dimensional linear programming that \textit{multiplicatively} considers all three matrices to decide the optimal replay set, resulting in an unacceptably complex computation.
Here, we propose an approximate algorithm that identifies local optimal data points within each matrix segment and \textit{additively} combines all three factors into consideration with significantly reduced computation.
Specifically, let the size of the replay set be $n_r$ for each domain, we first rearrange $\mathbf{F}^t$ in ascending order based on the magnitude distance $\mathbf{M}^t$. Then, we divide $\mathbf{F}^t$ into $ \frac{n_r}{2}$ equal-length segments $\mathbf{F}^t=\{\mathbf{F}^t_{1:\frac{2n}{n_r}}, 
\dots,\mathbf{F}^t_{(n-\frac{2n}{n_r}):n}\} \in \mathbb{R}^{\frac{n_r}{2} \times \frac{2n}{n_r}\times d}$.
Within \textit{each} segment, we identify the most stable feature \( \mathbf{f}_{s}^t \) based on \( \mathbf{S}^t \) and include its corresponding image \( \mathbf{x}_s^t \) into the replay set. Then, to simultaneously consider the uniformity of angularity (\textit{i.e.}, $\mathbf{A}^t$), we search for the feature within each segment that has the lowest normalized cosine similarity with $\mathbf{f}_{s}^t$ termed $\mathbf{f}_{a}^t$. Subsequently, we could select $\frac{n_r}{2}$ number of $\mathbf{f}_{s}^t$ and $\mathbf{f}_{a}^t$ from all segments. Their corresponding images are stored to constitute the $t$-th replay set of one domain (Real or Fake). We provide the concisely summarized algorithm of SUR in \textit{Supplementary Material}.

\subsection{Latent-space Incremental Detector (LID)}
We propose the Latent-space Incremental Detector (LID) to stack previous and new tasks brick by brick in the latent space. LID comprises two key elements: feature isolation and incremental decision alignment. 
\subsubsection{Feature Isolation with Distribution Re-filling}
Here, we seek to isolate the distributions of each real/fake and previous/new domain and mitigate override to preserve knowledge and accumulate the learned forgery information from both new and previous tasks.

\textbf{Distribution Re-filling (DR).} To further facilitate the isolation of different distributions, we propose leveraging the sparse uniformity of the SUR set to refill the latent-space distribution between replayed data points and centroids. Specifically, since SUR can be viewed as a uniform sparse subset of the previous global distribution, the space between SUR features and the centroids should also belong to the same previous global distribution. Therefore, we can employ latent space mixup to refill and further simulate the previous global distribution, aiding in enhanced feature isolation. The operation of the proposed distribution re-filling involves two random features ($\mathbf{f}_1$ and $\mathbf{f}_2$) from the same replay set and their corresponding centroid ($\mathbf{c}$). This can be formulated as:
\begin{equation}
    \mathbf{f}_{\text{filled}}=\beta(\alpha \mathbf{f}_1+(1-\alpha)\mathbf{f}_2)+(1-\beta) \mathbf{c},
\end{equation}
where $\alpha,\beta \in [0,1]$ are random mixing ratios. By doing so, we can effectively re-fill the triangular region formed by vertices $\mathbf{f}_1$, $\mathbf{f}_2$, and $\mathbf{c}$, further facilitating feature isolation when training on the new task.

\textbf{Isolation Loss.} With SUR and re-filled data, we can introduce supervised contrastive loss~\cite{scloss} to isolate each feature domain of real/fake and previous/new distributions. Formally, the isolation loss could be written as:
\begin{equation}
    \mathcal{L}_{iso} = -\frac{1}{N} \sum_{i=1}^{N} \log \left( \frac{\exp(\mathbf{f}_i \cdot \mathbf{f}_j / \tau)}{\sum_{k=1}^{N} \mathbb{I}_{[y_i \neq y_k]} \exp(\mathbf{f}_i \cdot \mathbf{f}_k / \tau)} \right),
\end{equation}
where $\mathbf{f}_i,\mathbf{f}_j$ are features from the same domains. $y_i$ is the domain label of $\mathbf{f}_i$, and it is allocated with a \textbf{unique} value to each real/fake and previous/new 
domain. $\mathbb{I}_{[y_i \neq y_k]}$ denotes an indicator function, which equals 1 if $y_i\neq y_k$, and 0 otherwise. Notably, $\mathbf{f}$ could be the feature of current training data if they are from the new task, and generated by SUR or re-filled data if they are from the previous tasks. Meanwhile, to encourage the learning of diverse real domains, the real data from different tasks is also assigned with different unique $y_i$. 

Feature isolation prevents the distribution override of the incrementing tasks with the previous ones, thereby mitigating catastrophic forgetting. Meanwhile, it encourages the backbone extractor to differentiate among the domains of each task, thus improving its sensitivity to various types of forgery information.
\subsubsection{Incremental Decision Alignment}
While feature isolation reduces the feature override and improves the backbone's sensitivity to forgery information, it remains challenging to derive the final binary detection outcomes from the task-wise isolated domains straightforwardly. Therefore, we propose Incremental Decision Alignment (IDA) to effectively leverage the accumulated forgery information from multi-class isolated features for the final binary detection outcome. 

IDA aims at aligning the decision boundaries of each isolated Real/Fake domain across all tasks. In this way, we can encourage feature isolation while simultaneously optimizing an aligned decision boundary to divide all real and fake domains for the final detection.
For alignment, it is first necessary to train and obtain the individual real/fake boundary for each task separately. Therefore, we first assign and maintain independent classifiers to deal with the real and fake samples from the same task. These classifiers can be treated as the decision boundaries for each task individually. 
The classifier for the $t$-th task is denoted by $\mathcal{C}^t(\ast;\mathbf{\theta}^t)$, where $\mathbf{\theta}^t$ is the parameter of $\mathcal{C}^t$. To ensure alignment across all tasks, it is sufficient to focus on aligning the incremented $\mathcal{C}^{t+1}(\ast;\mathbf{\theta}^{t+1})$ with the previous $\mathcal{C}^t(\ast;\mathbf{\theta}^t)$, which thereby recursively aligning all tasks. As the classifiers for dividing Real/Fake are linear layers, aligning the decision boundaries is identical to ensuring the \textit{angularity consistency} of the linear parameters. Hence, one optimization step of decision alignment for $\mathcal{C}^{t+1}(\ast;\mathbf{\theta}^{t+1})$ could be formally written as:
\begin{equation}
\theta^{t+1} \leftarrow \left\| \theta^{t+1} \right\|_2 \cdot \frac{ (1-\gamma) \tilde{\theta}^{t+1} +\gamma \tilde{\theta}^t }{ \left\| (1-\gamma) \tilde{\theta}^{t+1} + \gamma \tilde{\theta}^t \right\|_2 }
, \label{eq:align}
\end{equation}
where $\tilde{\theta}=\frac{\theta}{\|\theta\|_2}$ and $\gamma$ denotes the learning rate. During training on the $(t+1)$-th task, the classifier $\mathcal{C}^{t+1}$ is optimized following Eq.~\ref{eq:align} to be aligned with $\mathcal{C}^{t}$, while all previous classifiers are frozen to maintain the previous decision boundaries and their alignment.
\subsection{Training and Inference}
\textbf{Training.} During training on $(t+1)$-th task, the $1$-st to $t$-th replay sets and $(t+1)$-th training data will be combined together to $\mathbf{X}=\{\hat{\mathbf{X}}^1,\hat{\mathbf{X}}^2,...,\hat{\mathbf{X}}^t,\mathbf{X}^{t+1}\}$. Then, their features $\mathbf{F}=\{\hat{\mathbf{F}}^1,\hat{\mathbf{F}}^2,...,\hat{\mathbf{F}}^t,\mathbf{F}^{t+1}\}$ can be extracted by $\mathbf{F}=\mathcal{E}^{t+1}(\mathbf{X})$. Following~\cite{dfil}, we also maintain the previous-task learned information via knowledge distillation loss, which can be written as:
\begin{equation}
    \mathcal{L}_{dis}=\sum_{i=1}^{t}(\hat{\mathbf{F}}^i-\mathcal{E}^{t}(\hat{\mathbf{X}}^i))^2.
\end{equation}
Note that $\mathcal{E}^{t}$ is the frozen backbone extractor trained on the previous $t$-th task. Subsequently, we deploy isolation loss ($\mathcal{L}_{iso}$) with distribution re-filling to achieve feature isolation. Finally, the binary detection loss could be formulated as:
\begin{equation}
    \mathcal{L}_{det}=\sum_{i=1}^{t}\text{CE}(\mathcal{C}^i(\hat{\mathbf{F}}^i),\textbf{Y}^i)+\text{CE}(\mathcal{C}^{t+1}(\mathbf{F}^{t+1}),\textbf{Y}^{t+1}),
\end{equation}
where $\text{CE}$ represents the Cross-Entropy Loss, $\textbf{Y}^t$ is the binary detection labels for the $t$-th task.
Therefore, the overall loss function could be written as:
\begin{equation}
    \mathcal{L}_{\text{overall}}= \mathcal{L}_{iso}+\mu_1 \mathcal{L}_{dis}+\mu_2 \mathcal{L}_{det},
\end{equation}
where $\mu_1$ and $\mu_2$ are trade-off parameters. After optimizing $\mathcal{L}_{\text{overall}}$ via backpropagation, we apply Eq.~\ref{eq:align} to optimize the decision boundary for alignment.\\
\textbf{Inference.} During inference, the input image $\mathbf{x}$ is first processed to feature $\mathbf{f}$ by $\mathcal{E}$. Since the specific task of $\mathbf{x}$ is unknown during inference in real-world applications, we cannot determine the specific classifier for inference. Considering all classifiers have aligned decision boundaries, we apply their average detection result as the final inference outcome, which can be written as:
\begin{equation}
    y_{\text{infer}}=\sum_{i=1}^{t+1}\frac{\mathcal{C}^i(\mathbf{f})}{t+1}.
\end{equation}
\input{Table_main}
\section{Experimental Results}
\subsection{Experimental Settings}
\paragraph{Datasets.} In experiments, we employ a diverse collection including both classical and cutting-edge datasets with three fundamental face forgery categories, \textit{i.e.,} Face-Swapping (FS), Face-Reenactment (FR), and Entire Face Synthesis (EFS)~\cite{df40}. Specifically, we employ three classical FS datasets, that is, Celeb-DF-v2 (CDF)~\cite{Celeb-df}, DeepFake Detection Challenge Preview (DFDCP)~\cite{dfdc}, and DeepFakeDetection (DFD)~\cite{dfd}. FaceForensics++ ~\cite{FF++} is constructed by four forgery methods including both FS and FR, therefore it could be treated as a dataset with Hybrid forgery categories. Moreover, we further deploy datasets released in 2024 with more diverse forgery categories and techniques, that is, \{MCNet~\cite{mcnet}, BlendFace~\cite{blendface}, StyleGAN3~\cite{stylegan3}\} from DF40~\cite{df40} and \{SDv21~\cite{sdv15}\} from DiffusionFace~\cite{diffusionface}.
\paragraph{Incremental Protocols.} To systematically analyze the effectiveness of different approaches in incremental face forgery detection, we introduce three incremental protocols for evaluation.
\begin{itemize}
    \item \textbf{Protocol 1 (P1): Datasets Incremental} with \{SDv21, FF++, DFDCP, CDF\}. \\ Following the rapid development of new forgery datasets with three different categories (\textit{i.e.}, FS, FE in FF++, and EFS in SDv21), where both real and fake data are novel.
    \item \textbf{Protocol 2 (P2): Forgery Categories Incremental} with \{Hybrid (FF++), FR (MCNet), FS (BlendFace), EFS (StyleGAN3)\}. \\ Following the development of new forgery techniques in one specific real-world scenario, where real is the same while only fake data are novel and vary in categories.
    \item \textbf{Protocol 3 (P3):} \{FF++, DFDCP, DFD, CDF\}. \\ Classical protocol from previous works~\cite{dfil,dmp}.
\end{itemize}
\paragraph{Implementation Details.} 
For face preprocessing, we strictly follow the official code and settings provided by the standardized benchmark DeepFakeBench~\cite{deepfakebench}. 
Then, we carefully reproduce all baseline methods within the DeepFakeBench and employ the same training configuration to ensure a fair comparison.
EfficientNetB4~\cite{effnet} is employed as the backbone of our detector. 
The Adam optimizer is used with a learning rate of 0.0002, epoch of 20, input size of 256 $\times$ 256, and batch size of 32. The replay buffer size of each task is 500 for methods that require replaying (including HDP~\cite{hdp}). The trade-off parameters are set as $\mu_1=1$, $\mu_2=0.1$, and $\gamma=0.001$.
\textit{Frame-level} Area Under Curve (AUC)~\cite{deepfakebench} is applied as the major evaluation metric of experimental results. While accuracy (ACC) is also used to align the metric with existing methods~\cite{dfil,dmp}. All experiments are conducted on one NVIDIA Tesla A100 GPU.
\subsection{Comparisons with Existing Methods for Incremental Face Forgery Detection}
To comprehensively evaluate the IFFD performance, we compare our method with existing SoTA methods on P1 and P2. These comparing methods include classical general incremental learning methods (\textit{i.e.}, LwF~\cite{lwf}, iCaRL~\cite{icarl}, and DER~\cite{der}) and deepfake incremental learning methods (\textit{i.e.}, CoReD~\cite{cored}, DFIL~\cite{dfil}, and HDP~\cite{hdp}). They are carefully reproduced to be evaluated on P1 and P2 with the same experimental setting strictly based on their official code. As shown in Tab.~\ref{tab:main_P1P2}, the results substantially demonstrate the significant improvement of our method in both practical scenarios. Notably, the existing IFFD methods fail to perform promisingly in P2, where forgery methods are diverse and real images are in the same domain. In this scenario, the detectors are more prone to overriding previously learned information because forgery-irrelevant information is consistent across different forgeries, making their features more similar. This implies that they may not fully capture the specific forgery pattern and override the learned previous forgery information. 

In supplementary material, we provide results based on Protocol 3, which also indicates the superior performance of our method.

\subsection{Ablation Study}
Here, we evaluate the significance and effectiveness of each proposed component, that is, the Sprase Uniform Replay (SUR) strategy, Distribution Re-filling (DR), Isolation Loss ($\mathcal{L}_{iso}$), and Incremental Decision Alignment (IDA).
Notably, since the SUR strategy provides the previous global distribution that is indispensable to our overall framework, we particularly investigate it in the second paragraph. All presented results for the ablation study are trained after incrementing four datasets with Protocol 1.
\begin{table}[b]
\centering
\small
\centering
\begin{tabular}{lccccc}\toprule
Variant & SDv21 & FF++ & DFDCP & CDF & Avg. \\ \toprule
w/o All & 0.8731 & 0.6737 & 0.7685 & 0.9716 &  0.8217 \\
w/o IDA       &       0.8579&      0.7571&       0.7439&     0.9817&     0.8352\\ 
w/o  $\mathcal{L}_{iso}$     &       0.9597&      0.8012&       0.8239&     0.9517&      0.8841\\
w/o DR        &       0.9759&      0.8322&       0.8806&     \textbf{0.9754}&      0.9160\\ \midrule
Ours      &       \textbf{0.9997}&      \textbf{0.8479}&       \textbf{0.9067}&     0.9744&   
\textbf{0.9315}\\ \bottomrule
\end{tabular}
\caption{Ablation study (AUC) for each proposed component.}\label{tab:abl-main}
\end{table}
\begin{table}[b]
\centering
\centering
\footnotesize
\setlength{\tabcolsep}{4.5pt}
\begin{tabular}{lccccc}\toprule
Strategy & SDv21 & FF++ & DFDCP & CDF & Avg. \\ \toprule
Center        &       0.9011&      0.8106&       0.7431&     0.9817&      0.8591\\
Center+Hard      &       0.9501&      0.7969&       0.8103&     0.9531&      0.8776\\
Random        &       0.6954&      0.7538&       0.7395&     0.9417&      0.7826\\
Random Uniform     &       0.9677&      \textbf{0.8529}&       0.8445&     0.9624&      0.9069\\ \midrule
SUR (Ours)      &       \textbf{0.9971}&      0.8479&       \textbf{0.9067}&     \textbf{0.9744}&   
\textbf{0.9315}\\ \bottomrule
\end{tabular}
\caption{Ablation study (AUC) for different replay strategies.}~\label{tab:abl-replay}
\end{table}

\begin{figure}[t]
    \centering
    \includegraphics[width=1\linewidth]{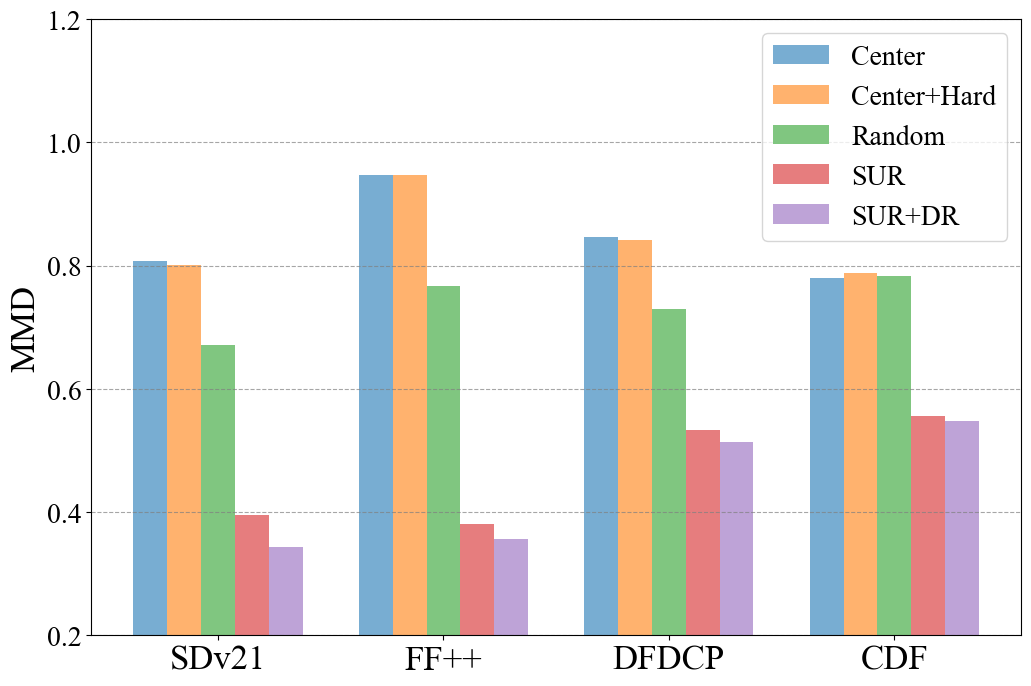}
    \caption{Evaluations of global distinction between the replay set and the training set. Maximum Mean Discrepancy (MMD) between different replay sets and their corresponding original training sets is deployed as the evaluation metric. A lower MMD indicates a smaller distinction between the replay set and the training set.}
    \vspace{-0.3cm}
    \label{fig:replay}
\end{figure}

\paragraph{Overall Ablation.}
As shown in Tab.~\ref{tab:abl-main}, we design ablation variants that remove each component respectively to assess their effectiveness. It can be observed that w/o IDA, the detector cannot leverage the accumulated forgery information and hence it exhibits poor performance. While $\mathcal{L}_{iso}$ also plays a crucial role in performance improvement. In addition, the proposed DR further enhances the IFFD performance of our method.
\paragraph{Effect of SUR Compared with Other Replay Strategies.}
To demonstrate the superiority of the proposed SUR strategy, we replace SUR with other existing replay strategies, that is, Center (C), Center+Hard (C+H), Random (R), and Random Uniform (RU). Specifically, C and C+H are following the implementation from DFIL~\cite{dfil}. R denotes randomly sampled from all training data. RU represents replacing ``choosing a stable subset'' with ``choosing a random subset'' from the uniform subsets. The results in Tab.~\ref{tab:abl-replay} show that the proposed uniform sampling strategy could significantly enhance the performance of aligned feature isolation, while considering the stability factor could further strengthen its effectiveness. 
Additionally, we evaluate the distribution distinctions between replay sets and their corresponding original training sets via Maximum Mean Discrepancy (MMD)~\cite{MMD}, which is a statistical method used to measure the distinction between two distributions.
As shown in Fig.~\ref{fig:replay}, existing methods ignore to maintain the global feature distributions, hence their MMDs are even larger than Random. In contrast, the proposed SUR can effectively simulate the global distribution of training tasks, and the proposed Distribution Refilling (DR) could enhance the performance of the simulation.

For sensitivity evaluations of our method about \textit{robustness against perturbations} and the \textit{size of replay set}, please refer to \textit{Supplementary Material}.
\begin{figure*}[htbp]
    \centering
    \includegraphics[width=1\linewidth]{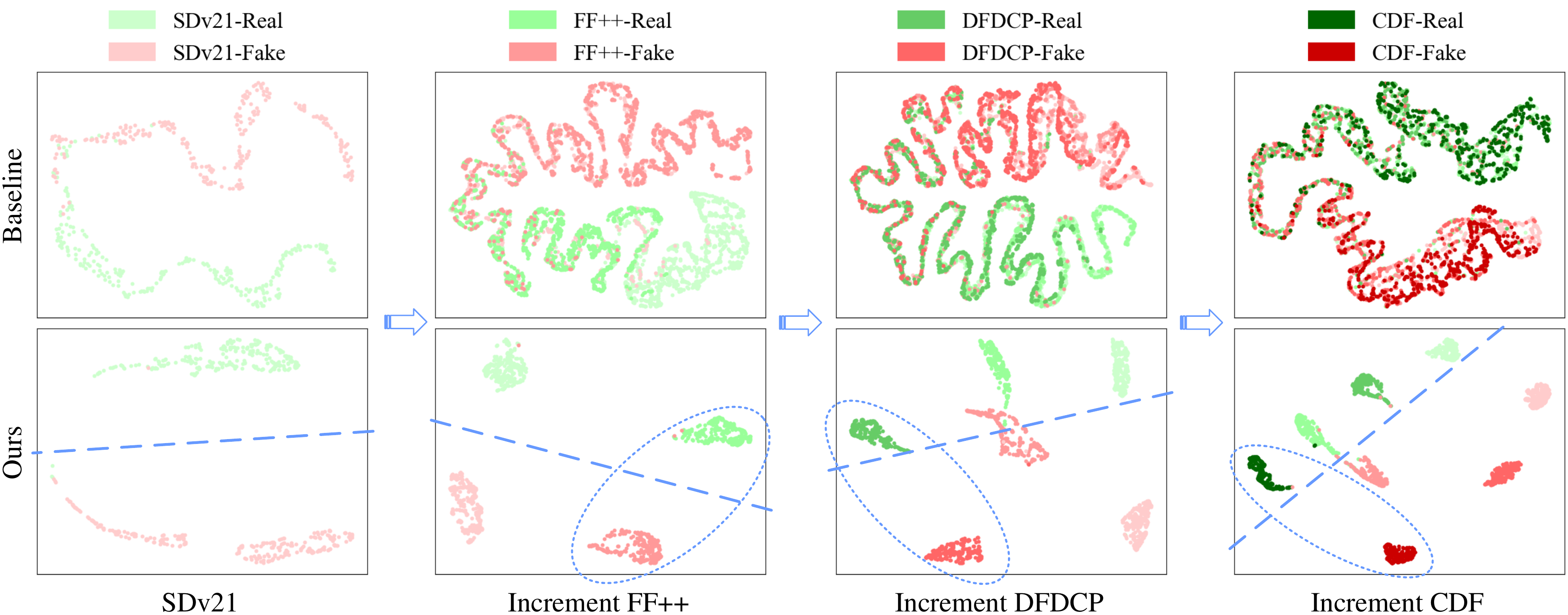}
    \caption{UMAP~\cite{umap} latent-space visualization for IFFD with Protocol 1. The upper row is the results of the baseline method (DFIL~\cite{dfil}) while the lower row is Ours. All shapes in blue are added for better illustration. The dashed lines denote the aligned boundary that divides real and fake. The dotted circles contain the distributions of newly incremented tasks. }
    \label{fig:Umap-dist}
\end{figure*}
\subsection{Visualization of Latent-Space Distribution}

Considering that the learned distribution of features is crucial to demonstrate the proposed aligned feature isolation, we carefully design experiments for the visualization of latent space distribution to investigate the effectiveness of our method. Here, we utilized UMAP~\cite{umap} to reduce feature dimension for visualizing the latent space distribution. As shown in Fig.~\ref{fig:Umap-dist}, we sequentially increment datasets following Protocol 1. We apply DFIL~\cite{dfil} as the comparison Baseline of our method. It can be observed that the Baseline continuously overrides the previous distribution with the incremented one, which leads to its severe forgetting issue and poor detection performance. In contrast, our method achieves incrementing new tasks with isolated distributions and aligned decision boundaries for the final binary detection.
More results for latent-space visualization can be found in \textit{Supplementary Material}.

\section{Conclusion}
In this paper, we propose the novel aligned feature isolation to improve the performance of Incremental Face Forgery Detection (IFFD). Specifically, we consider stacking the feature distributions of incrementing and previous tasks ``brick by brick'' to mitigate the global distribution overriding, accumulate diverse forgery information, and thus address the catastrophic forgetting issue. Subsequently, we propose a novel Sparse Uniform Replay (SUR) strategy and Latent-space Incremental Detector (LID) to realize aligned feature isolation. Experiments on a novel advanced IFFD evaluation benchmark substantially demonstrate the superiority of the proposed method.

\section*{Acknowledgments and Disclosure of Funding}
We would like to thank all the reviewers for their constructive comments. Our work was supported by the National Natural Science Foundation of China (NSFC) under Grant  No.62171324, No.62371350, and No.62372339.

{\small
\bibliographystyle{ieeenat_fullname}
	\bibliography{refer,refer_incremental} 
}
\clearpage
\section*{Supplementary Materials}  

    
\setcounter{section}{0}  
\input{body}
\end{document}

%% file: Table_main.tex
\begin{table*}[]
\centering
\centering
\footnotesize
\begin{tabular}{l@{\hspace{3pt}}cc|ccccc|ccccc} \toprule
\multirow{2}{*}{Method}                                               & \multirow{2}{*}{Replays}&\multirow{2}{*}{Task}  
& \multicolumn{5}{c|}{Protocol 1}             & \multicolumn{5}{c}{Protocol 2}             \\ \cmidrule(lr){4-8} \cmidrule(lr){9-13}
                                                                      & 
&
& SDv21  & FF++   & DFDCP  & CDF    & Avg.   & Hybrid & FR     & FS     & EFS    & Avg.   \\ \toprule
\multirow{4}{*}{Lower Bound} & \multirow{4}{*}{0}&T1                     
& 0.9998 &        -&        -&        -&        0.9998& 0.9687 &        -&        -&        -&        0.9687\\
                                                                      & 
&T2                     
& 0.7392 & 0.9460 &        -&        -&        0.8426& 0.5037 & 0.9999 &        -&        -&        0.7685\\
                                                                      & 
&T3                     
& 0.7004 & 0.7136 & 0.9133 &        -&        0.7758& 0.5790 & 0.1497 & 0.9956 &        -&        0.5748\\
                                                                      & 
&T4                     
& 0.5280 & 0.6362 & 0.7636 & 0.9816 & 0.7260 & 0.5078 & 0.6261 & 0.4834 & 1.0000 & 0.6536 \\ \midrule
\multirow{4}{*}{LwF~\cite{lwf} (TPAMI'17)}& \multirow{4}{*}{0}&T1                     
& 0.9998 &        -&        -&        -&        0.9998& 0.9700 &        -&        -&        -&        0.9700\\
                                                                      & 
&T2                     
& 0.7532 & 0.9479 &        -&        -&        0.8506& 0.8876 & 0.8845 &        -&        -& \textbf{0.8861}\\
                                                                      & 
&T3                     
& 0.5807 & 0.9050 & 0.8370 &        -&        0.7742& 0.8407 & 0.8099 & 0.9644 &        -& \underline{0.8724}\\
                                                                      & 
&T4                     
& 0.6154 & 0.8133 & 0.8336 & 0.9263 & 0.7972 & 0.7873 & 0.5673 & 0.9367 & 0.9282 & 0.8049 \\ \midrule
\multirow{4}{*}{iCaRL~\cite{icarl} (CVPR'17)}& \multirow{4}{*}{500}&T1                     
& 0.9998 &        -&        -&        -&        0.9998& 0.9653 &        -&        -&        -&        0.9653\\
                                                                      & 
&T2                     
& 0.9267 & 0.9479 &        -&        -& 0.8363 & 0.6736 & 0.9989 &        -&        -& 0.8363 \\
                                                                      & 
&T3                     
& 0.9010 & 0.7447 & 0.9135 &        -&        0.7864& 0.7379 & 0.6624 & 0.9754 &        -&        0.7919\\
                                                                      & 
&T4                     
& 0.8520 & 0.6789 & 0.7501 & 0.9805 & 0.8154 & 0.5298 & 0.5538 & 0.6474 & 1.0000 & 0.6828 \\ \midrule
\multirow{4}{*}{DER~\cite{der} (CVPR'21)}& \multirow{4}{*}{500}&T1                     
& 0.9998 &        -&        -&        -&        0.9998& 0.9700 &        -&        -&        -&        0.9700\\
                                                                      & 
&T2                     
& 0.7353 & 0.9518 &        -&        -&        0.8436& 0.5903 & 0.9973 &        -&        -& 0.7938 \\
                                                                      & 
&T3                     
& 0.6378 & 0.7402 & 0.9092 &        -&        0.7624& 0.6815 & 0.1968 & 0.9794 &        -& 0.6193 \\
                                                                      & 
&T4                     
& 0.6071 & 0.6560 & 0.7654 & 0.9873 & 0.7539 & 0.5679 & 0.5983 & 0.6536 & 1.0000 & 0.7049 \\ \midrule
\multirow{4}{*}{CoReD~\cite{cored} (MM'21)}& \multirow{4}{*}{500}&T1                     
& 0.9998 &        -&        -&        -&        0.9998& 0.9665 &        -&        -&        -&        0.9665\\
                                                                      & 
&T2                     
& 0.7459 & 0.9433 &        -&        -&        0.8446& 0.9355 & 0.7988 &        -&        -& 0.8671 \\
                                                                      & 
&T3                     
& 0.8555 & 0.9096 & 0.8154 &        -&        0.8602& 0.8907 & 0.7929 & 0.8605 &        -& 0.8480 \\
                                                                      & 
&T4                     
& 0.8718 & 0.8376 & 0.7987 & 0.9341 &        0.8606& 0.8454 & 0.6429 & 0.8417 & 0.9263 & \underline{0.8141}\\ \midrule
\multirow{4}{*}{DFIL~\cite{dfil} (MM'23)}& \multirow{4}{*}{500}&T1                     
& 0.9998 &        -&        -&        -&        0.9998& 0.9646 &        -&        -&        -&        0.9646\\
                                                                      & 
&T2                     
& 0.7400 & 0.9466 &        -&        -&        0.8433& 0.5574 & 0.9975 &        -&        -& 0.7775 \\
                                                                      & 
&T3                     
& 0.9692 & 0.8164 & 0.9088 &        -&        \underline{0.8981}& 0.6071 & 0.6649 & 0.9903 &        -& 0.7541 \\
                                                                      & 
&T4                     
& 0.9326 & 0.7397 & 0.7908 & 0.9881 & 0.8628 & 0.5083 & 0.9556 & 0.7081 & 0.9996 & 0.7929 \\ \midrule
\multirow{4}{*}{HDP~\cite{hdp} (IJCV'24)}& \multirow{4}{*}{500}&T1                     
&        0.9998&        -&        -&        -&        0.9998&        0.9671&        -&        -&        -&        0.9671\\
                                                                      & 
&T2                     
&        0.8373&        0.9507&        -&        -&        \underline{0.8940}&        0.6741&        0.9545&        -&        -&        0.8143\\
                                                                      & 
&T3                     
&        0.9341&        0.8532&        0.8737&        -&        0.8870&        0.6300&        0.7135&        0.9509&        -&        0.7648\\
                                                                      & 
&T4                     
&        0.9055&        0.8039&        0.8412&        0.9501&        \underline{0.8752}&        0.5989&        0.7006&        0.8934&        0.9373&        0.7826\\ \midrule
\multirow{4}{*}{\textbf{\textit{SUR-LID (Ours)}}}                                                 & \multirow{4}{*}{500}&T1                     
& 0.9999 &        -&        -&        -&        0.9999& 0.9685 &        -&        -&        -&        0.9685\\
                                                                      & 
&T2                     
& 0.9937 & 0.9485 &        -&        -& \textbf{0.9711}& 0.8291 & 0.9242 &        -&        -& \underline{0.8766}\\
                                                                      & 
&T3                     
& 0.9986 & 0.8844 & 0.9161 &        -& \textbf{0.9330}& 0.9050 & 0.9626 & 0.9794 &        -& \textbf{0.9490}\\
                                                                      & &T4                     & 0.9971 & 0.8479 & 0.9067 & 0.9744 & \textbf{0.9315}& 0.8790 & 0.9679 & 0.9356 & 0.9907 & \textbf{0.9433}\\ \bottomrule
\end{tabular}
\caption{Performance comparisons (AUC) with Protocol 1 (Dataset Incremental) and Protocol 2 (Forgery Type Incremental). Lower Bound denotes vanilla incremental learning without any strategy. Task 1 (T1) to Task 4 (T4) represent current incremented tasks in \{SDv21, FF++, DFDCP, CDF\} or \{Hybrid, FR, FS, EFS\}. The underline represents the second best results while the bold denotes the best ones.}\label{tab:main_P1P2}
\end{table*}

%% file: body.tex
\section{Further Results Comparing with SoTA}
\subsection{Results with Protocol 3}
\begin{table}[htbp]
\centering
\small
\centering
\begin{tabular}{lccccc}\toprule
Method & FF++ & DFDCP & DFD & CDF & Avg. \\ \midrule
LwF~\cite{lwf}    &      67.34&       67.43&     84.05&     87.90&      76.68\\
CoReD~\cite{cored}  &      74.08&       76.59&     93.41&     80.78&      81.22\\
DFIL~\cite{dfil}   &      86.28&       79.53&     92.36&     83.81&      85.49\\
DMP~\cite{dmp}    &      \textbf{91.61}&       84.86&     91.81&     91.67&      89.99\\ \midrule
Ours   &      90.89&      \textbf{ 89.33}&     \textbf{93.97}&     \textbf{94.34}&      \textbf{92.13}\\ \bottomrule
\end{tabular}
\caption{Performance comparisons (ACC) with Protocol 3. All results of previous methods are copied from ~\cite{dmp} and ~\cite{dfil}.}\label{tab:P3}
\end{table}

In Tab.~\ref{tab:P3}, we copy the results after all tasks are incremented with P3 from their official papers~\cite{dfil,dmp} to further compare the IFFD performance. Despite the notable distinction in experimental settings among these methods, our method still exhibits superior performance.

\begin{table}
\centering
\begin{tabular}{lcccc}\toprule
Method & SDv21 & FF++  & DFDCP & Avg.  \\\toprule
Lower Bound &47.19&32.75&16.40&32.11\\
LwF &38.45&14.20&0.41&17.69\\
CoReD &12.80&11.21&2.05&8.69\\
DFIL    & 6.72  & 20.69 & 11.80 & 13.07 \\
HDP     & 9.43  & 14.68 & 3.25  & 9.12  \\ \midrule
Ours    & \textbf{0.28}  & \textbf{10.06} & \textbf{0.94}  & \textbf{3.09}   \\ \bottomrule
\end{tabular}
\caption{Evaluation of Forgetting Rate ${\downarrow}$ (\%).}\label{tab:fr}
\end{table}
\subsection{Evaluation with Forgetting Rate}
Following~\cite{liu2020mnemonics}, we compute FR based on AUC between current and first-learned models. Specifically, FR is calculated as $FR=1-\frac{AUC_{last}}{AUC_{first}}$, where $AUC_{last}$ is the AUC of one dataset tested on the currently-trained model, $AUC_{first}$ is the AUC of the model that firstly-introduced the dataset. The FR results in Tab.~\ref{tab:fr} indicate that our method has effectively tackled the issue of forgetting.

\section{Further Visualization Analysis}
\subsection{Visualization of Model Attention via Grad-CAM}
\begin{figure}[htbp]
    \centering
    \includegraphics[width=1\linewidth]{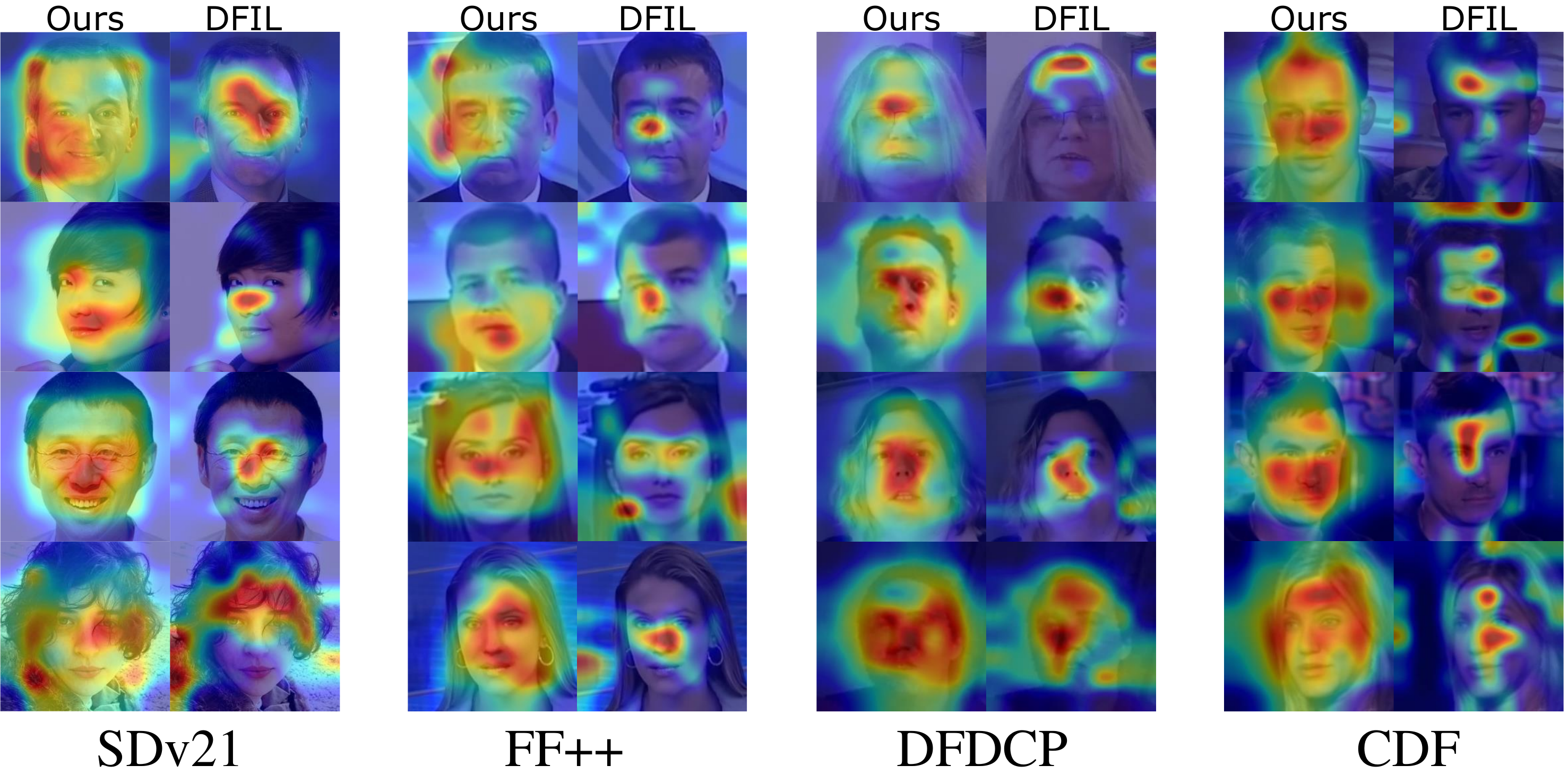}
    \caption{Saliency map visualization of DFIL~\cite{dfil} and the proposed method.}
    \label{fig:gradcam}
\end{figure}

As shown in Fig~\ref{fig:gradcam}, we deploy Grad-CAM~\cite{gradcam} to generate saliency maps. It can be observed that our method could explore more forgery clues since we successfully accumulated forgery information. While DFIL struggles to find rich clues and cannot consistently focus on the forgery regions. 
\subsection{Visualization of Actual Feature Distribution with Toy Models}
\begin{figure}[htbp]
    \centering
    \includegraphics[width=1\linewidth]{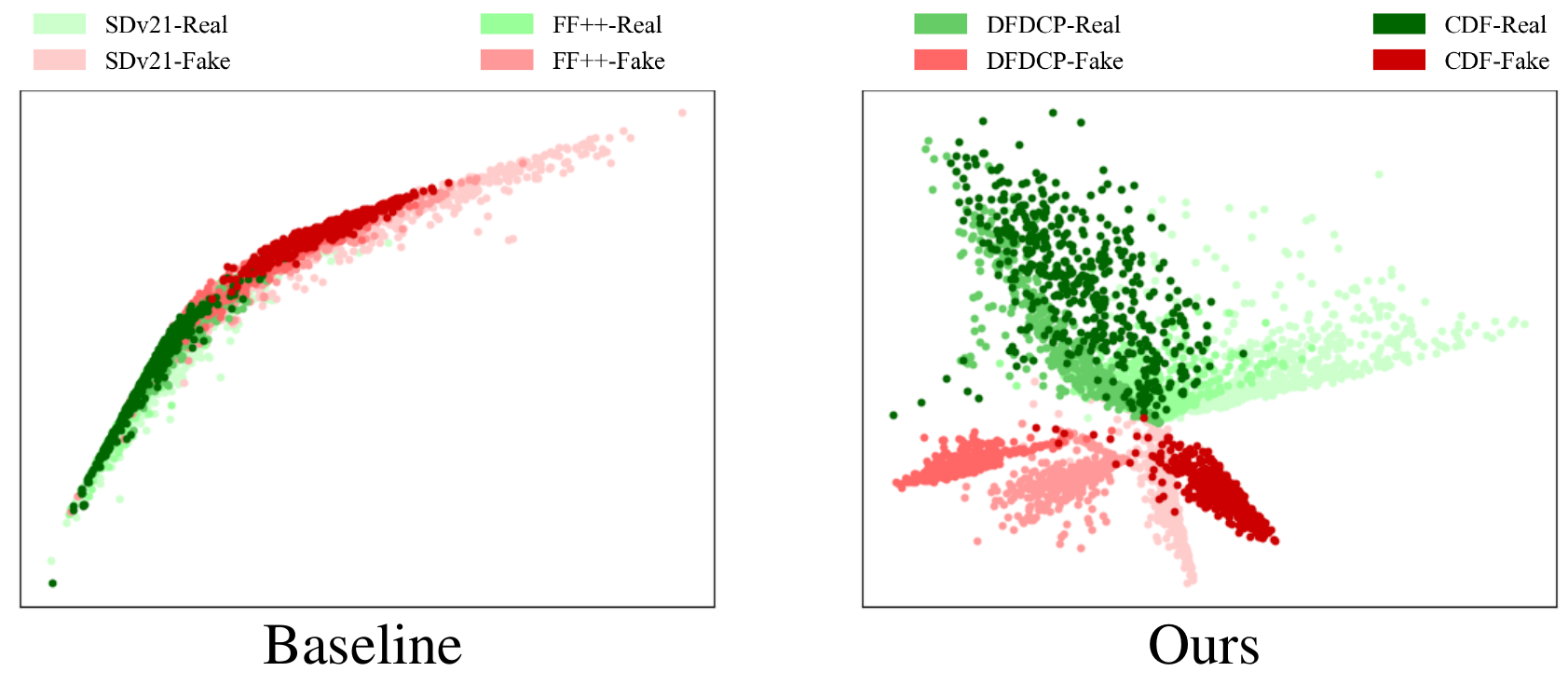}
    \caption{\textbf{Actual} two-dimensional feature distributions of toy models with Protocol 1.}
    \label{fig:Toy-dist}
\end{figure}

To further investigate the learned feature distribution in IFFD, we cleverly craft toy models to visualize the \textbf{actual} feature distributions of baseline (DFIL~\cite{dfil}) and our method. To be specific, we train new models with features that have only two dimensions and all other settings are consistent with the standard ones. Consequently, we could directly visualize the two-dimensional features with a two-dimensional coordinate system. As shown in Fig.~\ref{fig:Toy-dist}, the Baseline performs limited in distinguishing various forgeries and detecting binary Real/Fake, while our method could effectively isolate each domain and uphold a clean binary decision boundary. Notably, the two-dimensional features are insufficient to adequately represent the learned representations, resulting in the toy model performing poorly compared to the standard model. Nevertheless, it could still suggest that the actual feature distribution of the standard models is organized as we anticipated, that is, aligned feature isolation.
\begin{table*}[htb]
\centering
\begin{tabular}{lccccc}\toprule
Method     & DFD~\cite{dfd}           & UniFace~\cite{uniface}       & SDv15~\cite{sdv15}     & FakeAVCeleb~\cite{fakeavceleb}   & Avg.           \\ \midrule
Lower Bond & 0.6705 / 0.7038 & 0.6058 / 0.6216 & 0.5319 / -  & 0.5841 / 0.5995 & 0.5981 / 0.6416 \\
DFIL~\cite{dfil}       & 0.7719 / 0.8293 & 0.5637 / 0.6001 & 0.7786 / -  & 0.6111 / 0.6306 & 0.6813 / 0.6867 \\
HDP~\cite{hdp}        & 0.8039 / 0.8441 & 0.5971 / 0.6714 & 0.7211 / -  & 0.6535 / 0.6917 & 0.6939 / 0.7357 \\ \midrule
Ours       & \textbf{0.8225} / \textbf{0.8803} & \textbf{0.7269} / \textbf{0.7667} & \textbf{0.8110} / -  & \textbf{0.7663} / \textbf{0.8304} & \textbf{0.7817} / \textbf{0.8258} \\ \bottomrule
\end{tabular}
\centering
\caption{Cross-dataset evaluations for generality with \textit{frame-level} / \textit{video-level} AUC. SDv15 has no video-level result since it is an image-level dataset. All methods are trained based on Protocol 1 (SDv21, FF++, DFDCP, CDF) and tested on other unseen datasets. The best results are highlighted in \textbf{bold}.}\label{tab:gene}
\end{table*}

\section{Experiments of Generalization Ability}
\subsection{Generalization to Other Unseen Datasets}
To validate that the accumulated forgery information enables our method to learn more about forgery generality, we conduct cross-dataset experiments for generalization ability evaluation. As shown in Tab.~\ref{tab:gene}, we apply the model trained on Protocol 1 to be evaluated on DeepFakeDetection (DFD)~\cite{dfd}, UniFace~\cite{uniface} from DF40~\cite{df40}, SDv15 from DiffusionFace~\cite{diffusionface}, and FakeAVCeleb~\cite{fakeavceleb}. The experimental results substantially demonstrate that our method exhibits superior generalization ability attributable to the accumulated forgery information during incremental learning.
\subsection{Generalization to Other Backbone}
We additionally deployed our method on two mainstream backbones (ResNet and Xception) and compared the results with those of the original backbones under the same replay size. As shown in Tab.~\ref{tab:backbone}, our method also significantly improves the performance of these backbones.
\begin{table*}[h]
\centering

\begin{tabular}{lccccc}\toprule
Method        & SDv21  & FF++  & DFDCP  & CDF    & Avg.   \\\toprule
Xception+Ours & 0.996$^{\uparrow65.8\%}$ & 0.767$^{\uparrow24.9\%}$ & 0.852$^{\uparrow13.6\%}$ & 0.951$^{\uparrow0.75\%}$ & 0.892$^{\uparrow22.6\%}$ \\
ResNet+Ours   & 0.993$^{\uparrow85.8\%}$ & 0.688$^{\uparrow16.0\%}$ & 0.861$^{\uparrow20.0\%}$ & 0.935$^{\uparrow 0.73\%}$ & 0.869$^{\uparrow25.4\%}$ \\
\bottomrule
\end{tabular}
\caption{Generalization to other backbones (AUC). $\uparrow$ denotes the improvement compared with vanilla backbones.}\label{tab:backbone}
\end{table*}

\begin{algorithm}[h]
\caption{Sparse Uniform Replay (SUR)}\label{alg:sur}
		\KwIn{$t$-th Dataset: $\mathbf{X}^t_{all} = \{\mathbf{X}^t_{real},\mathbf{X}^t_{fake}\}$;
        Feature Extractor Trained on $t$-th Dataset: $\mathcal{E}^t$;
        Replay size: $n_r$.}

        Initialize the $t$-th replay set $\mathbf{X}^t_{replay}$ as empty;
        
        \For{  $\mathbf{X}^t \sim \mathbf{X}^t_{all}$}{
        
        extract features of $\mathbf{X}^t$
        
        $\mathbf{F}^t=\mathcal{E}(\mathbf{X}^t)$
        
        calculate feature centroid
        
        $\mathbf{c}^t=avg(\mathbf{F}^t)$

        calculate magnitude matrix from $\mathbf{F}^t$ to $\mathbf{c}^t$

        $\mathbf{M}^t = \|\mathbf{F}^t - \mathbf{c}^t\|_2$
        
        calculate angularity matrix from $\mathbf{F}^t$ to $\mathbf{c}^t$

        $\mathbf{A}^t = \frac{(\mathbf{F}^t - \mathbf{c}^t)}{\|\mathbf{F}^t - \mathbf{c}^t\|_2}$

        rearrange $\mathbf{F}^t$ in ascending order based on $\mathbf{M}^t$

        divide $\mathbf{F}^t$ into $\frac{n_r}{2}$ equal-length segments

        $\mathbf{F}^t=\{\mathbf{F}^t_{1:\frac{2n}{n_r}}, 
\dots,\mathbf{F}^t_{(n-\frac{2n}{n_r}):n}\}$

        \For{  $\mathbf{F}^t_{seg} \sim \{\mathbf{F}^t_{1:\frac{2n}{n_r}}, 
\dots,\mathbf{F}^t_{(n-\frac{2n}{n_r}):n}\}$}{
        
            calculate similarity of each feature $\mathbf{f}_i^t$ in $\mathbf{F}^t_{seg}$ with its shuffled $\tilde{\mathbf{f}}_i^t$ as stability score 
            
            $    s_i^t = \frac{\tilde{\mathbf{f}}_i^t \cdot (\mathbf{f}_i^t)^\text{T}}{\|\tilde{\mathbf{f}}_i^t\|_2 \cdot \|\mathbf{f}_i^t\|_2}$

            store the $\mathbf{x}_m^t$ corresponding to $\mathbf{f}_m^t$ with largest $s_m^t$ into $\mathbf{X}^t_{replay}$

            calculate angularity similarity of each feature  $\mathbf{f}_j^t$ in $\mathbf{F}^t_{seg}$ with $\mathbf{f}_m^t$ based on $\mathbf{A}^t$

            store the $\mathbf{x}_a^t$ corresponding to $\mathbf{f}_a^t$ with largest angularity similarity into $\mathbf{X}^t_{replay}$
            
        }
        }
		\KwOut{$t$-th replay set $\mathbf{X}^t_{replay}$.}  
\end{algorithm}
\section{Algorithm for Sparse Uniform Replay}
As shown in Algorithm~\ref{alg:sur}, we provide a concisely summarized algorithm for better comprehension in the detailed implementation of the proposed sparse uniform replay (SUR).
\section{Sensitivity Evaluation}
\begin{figure}[htbp]
    \centering
    \includegraphics[width=1\linewidth]{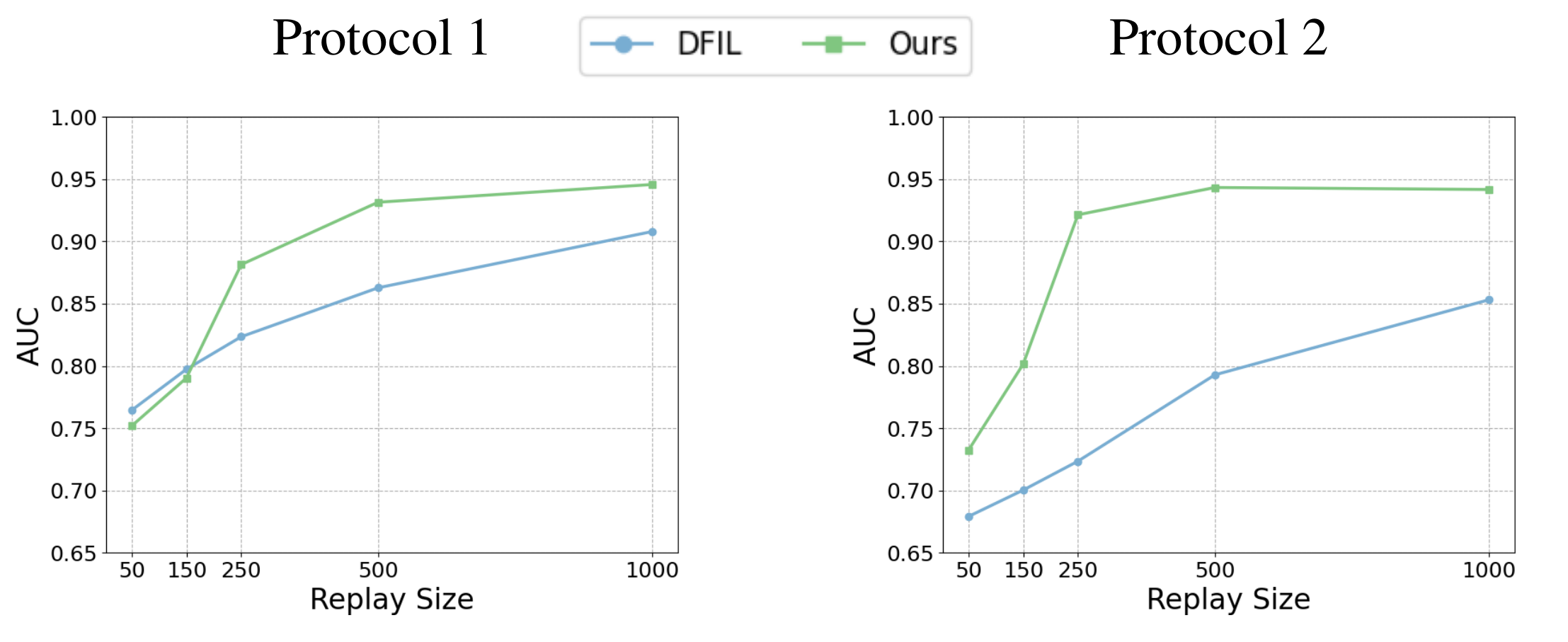}
    \caption{Sensitivity of replay size. The shown AUCs are the average values on four datasets after training with Protocol 1 or 2. }
    \label{fig:replay-size}
\end{figure}
\begin{figure*}[htbp]
    \centering
    \includegraphics[width=1\linewidth]{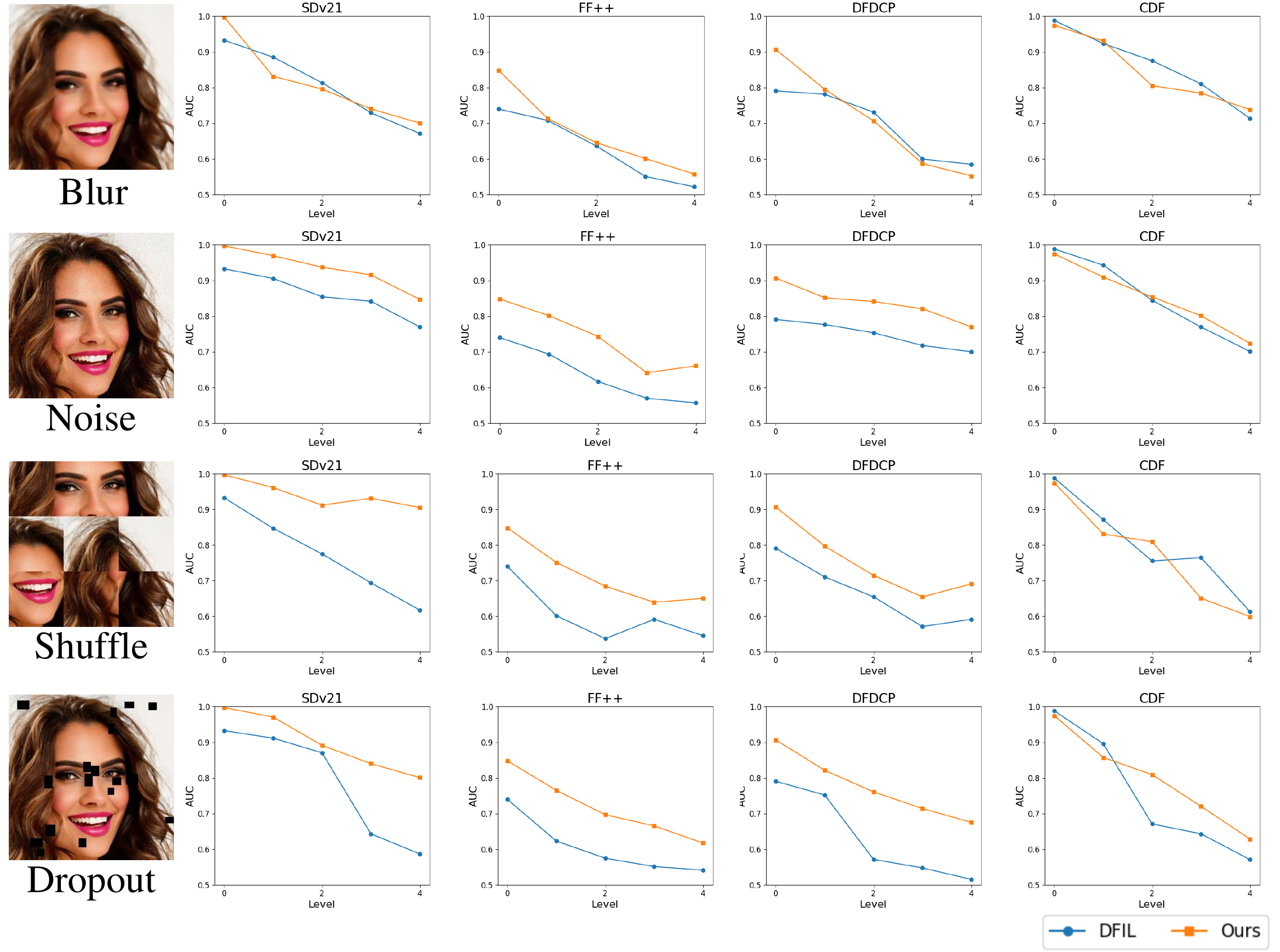}
    \caption{Robustness evaluations. The images in the first column are visualized illustrations of different types of applied perturbations. The models are trained based on Protocol 1.}
    \label{fig:robust}
\end{figure*}
\subsection{Effect of Replay Size}
In Fig.~\ref{fig:replay-size}, we examine the effect of the replay set size on model performance. It can be observed that the impact of replay set size on DFIL is relatively smooth, with performance gradually improving as the set size increases. In contrast, our method exhibits limited performance when the replay set size is small (\textit{i.e.}, 50, 150). This is because the constraints employed for the proposed aligned feature isolation rely heavily on the replayed global distribution. Nonetheless, once the replay set reaches a more standard size, the performance of our approach becomes superior and promising.

\subsection{Robustness against Unseen Perturbations}
Considering the importance of robustness for real-world applications, we evaluate the robustness of different IFFD methods against unseen perturbations. Specifically, based on Protocol 1, we assess robustness against Block-wise Dropout (Dropout), Grid Shuffle (Shuffle), Gaussian Noise (Noise), and Median Blur (Blur), each applied at multiple intensity levels. As shown in Fig.~\ref{fig:robust}, our method demonstrates consistent superiority in Noise, Shuffle, and Dropout, and also being comparable in Blur. The robustness superiority of our method may be attributed to the effective accumulation and utilization of forgery information achieved by our method, which enables the extracted and organized latent space to be more stable and representative.
